\documentclass{article}

\usepackage{arxiv}
\usepackage{algorithm}
\usepackage[noend]{algpseudocode}
\usepackage{wrapfig}
\usepackage{graphicx}
\usepackage{subfigure}
\usepackage{amsmath}

\newcommand{\B}{\fontseries{b}\selectfont}

\usepackage[utf8]{inputenc} 
\usepackage[T1]{fontenc}    
\usepackage{hyperref}       
\usepackage{url}            
\usepackage{booktabs}       
\usepackage{amsfonts}       
\usepackage{nicefrac}       
\usepackage{microtype}      
\usepackage{lipsum}

\title{Dynamic Neural Network Channel Execution for Efficient Training}

\author{
  Simeon E.~Spasov\thanks{Corresponding author} \\
  Department of Computer Science and Technology\\
  University of Cambridge\\
  Cambridge, 15 JJ Thomson Avenue,  \\
  \texttt{ses88@cam.ac.uk} \\
   \And
 Pietro ~Li\`o\\
   Department of Computer Science and Technology\\
  University of Cambridge\\
  Cambridge, 15 JJ Thomson Avenue,  \\
  \texttt{pl219@cam.ac.uk} \\
}

\begin{document}
\maketitle

\begin{abstract}
Existing methods for reducing the computational burden of neural networks at run-time, such as parameter pruning or dynamic computational path selection, focus solely on improving computational efficiency during inference. On the other hand, in this work, we propose a novel method which reduces the memory footprint and number of computing operations required for training \textit{and} inference. Our framework efficiently integrates pruning as part of the training procedure by exploring and tracking the relative importance of convolutional channels. At each training step, we select only a subset of highly salient channels to execute according to the combinatorial upper confidence bound algorithm, and run a forward and backward pass only on these activated channels, hence learning their parameters. Consequently, we enable the efficient discovery of compact models. We validate our approach empirically on state-of-the-art CNNs - VGGNet, ResNet and DenseNet, and on several image classification datasets. Results demonstrate our framework for dynamic channel execution reduces computational cost up to $4\times$ and parameter count up to $9\times$, thus reducing the memory and computational demands for discovering and training compact neural network models.
\end{abstract}

\keywords{neural networks \and channel execution \and channel selection \and dynamic training}

\section{Introduction}
The performance of convolutional neural networks (CNNs) in a variety of computer vision tasks, such as image classification, object detection and segmentation, has improved significantly and has surpassed other traditional methods. In recent years, network architectures have become deeper and more complex, yielding highly overparametrized models with a higher memory footprint and many floating-point operations. These requirements have restricted the application of CNNs on resource-constrained devices. Therefore, many techniques, such as pruning~\cite{han_weights}, which eliminate unnecesary parameters at run-time have been studied. These current methods have solely focused on making inference more cost-efficient.

\begin{wrapfigure}{r}{0.5\textwidth}
  \begin{center}
    \includegraphics[width=0.5\textwidth]{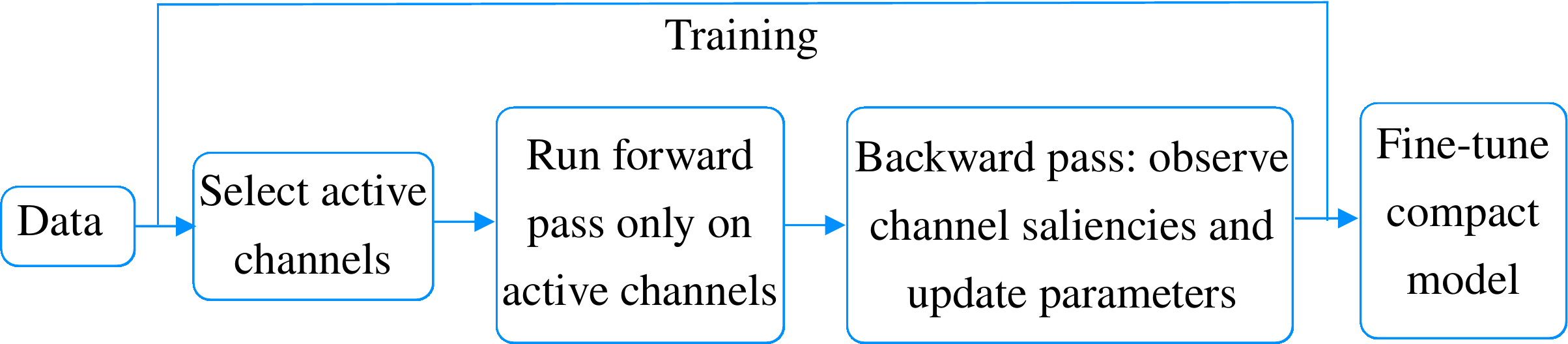}
  \end{center}
  \label{fig:flowchart}
  \caption{Flowchart of the dynamic channel execution framework for efficient training.}
\end{wrapfigure}

In this work, we investigate how to lower the memory and computational demands of CNNs during \textit{training} as well as inference. Figure~\ref{fig:flowchart} provides an overview of our methodology. We draw inspiration from dynamic neural networks~\cite{michigan_dnn} and selectively activate and execute a subset of all convolutional channels at each training step. We use a saliency criteria introduced by~\cite{Molchanov} to track the relatively contribution of executed channels to the network performance. Similarly to~\cite{Zhuang}, in our work we assume that a salient channel ought to possess discriminative power regardless of its position in the network. Consequently, we largely disregard channel position and connectivity, and assess each convolutional channel's importance independently of the others. Before each training step we employ a reinforcement learning algorithm to select a subset of channels to activate based on the saliency information we have gathered from the beginning of the training process. We then run a forward and backward pass only on these activated channels, hence executing them and learning their parameters. After identifying the most salient convolutional channels, we fine-tune their weights after freezing the network topology.

Our main contributions are summarized as follows. Firstly, we propose a dynamic channel execution methodology which enables training compact models directly by integrating the channel pruning procedure in the training process. In principle, the computational graph of the compact model in each training step comprises only active channels, therefore the memory footprint and computational cost of training are reduced. Our framework also gives direct control over the percentage of channels we want to activate and execute. Secondly, we formulate the channel selection problem as a combinatorial multi-armed bandit problem, and propose to use the combinatorial upper confidence bound algorithm (CUCB)~\cite{CUCB} to solve it. An advantage of this approach is that we do not require additional model complexity to implement the channel selection procedure unlike~\cite{michigan_dnn}. Experiments demonstrate we can significantly reduce the cost of training neural networks. For instance, a ResNet-50 network trained on the SVHN dataset with $30\%$ active channels outperforms the baseline with over $4\times$ parameter count reduction and $3\times$ floating-point operations reduction.  

\subsection{Related work}
\textbf{Adaptive computation} aims to change the network topology dynamically at run-time. The main focus has been reducing computational cost at inference~\cite{Veit2018, BlockDrop, skip_net, RNP_nips}. More recently,~\cite{google_gaternet} has investigated input-dependent dynamic filter selection. Their framework uses a backbone network, which performs the actual prediction, and a gater network, which decides which filters in the backbone to use. Sparsity in the number of gates, or active filters, is encouraged via L1 regularization. Y. Su and S. Zhou \emph{et al.} ~\cite{sensetime_dynamic_net} propose a dynamic inference method which provides various inference path options. This is achieved by dividing the original network models in blocks and utilizing a gating mechanism to predict the on/off status of each block during inference. Both of these recent methodologies introduce additional model complexity to achieve gating, and are not designed for efficient training on resource-constrained applications but rather only for inference. The closest work to ours is Dynamic Deep Neural Networks~\cite{michigan_dnn} which allows selective execution by integrating backpropagation with reinforcement learning. The proposed framework uses Q-learning to learn the parameters of control nodes that influence the computational path. The similarity of this setup to our work is that control nodes, and therefore certain network modules, can be on or off during training as well as inference, however, Dynamic Deep Neural Networks also require extra parameters for the control nodes, and the method is input-dependent.\par\vspace{1em}

\textbf{Pruning} is applied after training an initial network in order to eliminate model parameters, therefore making inference less computationally intensive, and decreasing model size and memory footprint. One approach is non-structured pruning which dates back to Optimal Brain Damage~\cite{opt_brain_dmg}. More recently, Han \emph{et al.}~\cite{han_weights} propose to prune individual weights with small magnitude, and Srinivas and Babu ~\cite{babu} propose to remove redundant neurons iteratively. The issue with non-structured pruning is that it requires specialized hardware ~\cite{han}. Structured pruning ~\cite{Zhuang, Chin_pruning, Molchanov, bmvc_pruning, net_slimming, Li_pruning} ,on the other hand, does not require dedicated libraries/hardware as it prunes whole filters, channels of convolutional kernels or even layers, based on some importance criteria. Other post-processing techniques to achieve compact network models include knowledge distillation~\cite{knowledge_distillation, Romero15-iclr}, weight quantization~\cite{xnor_net, binarized_nns}, low-rank approximation of weights~\cite{Lebedev, Emily_Denton}.

\section{Methodology}

\begin{figure}[!ht]
    \centering
    \includegraphics[width=\textwidth]{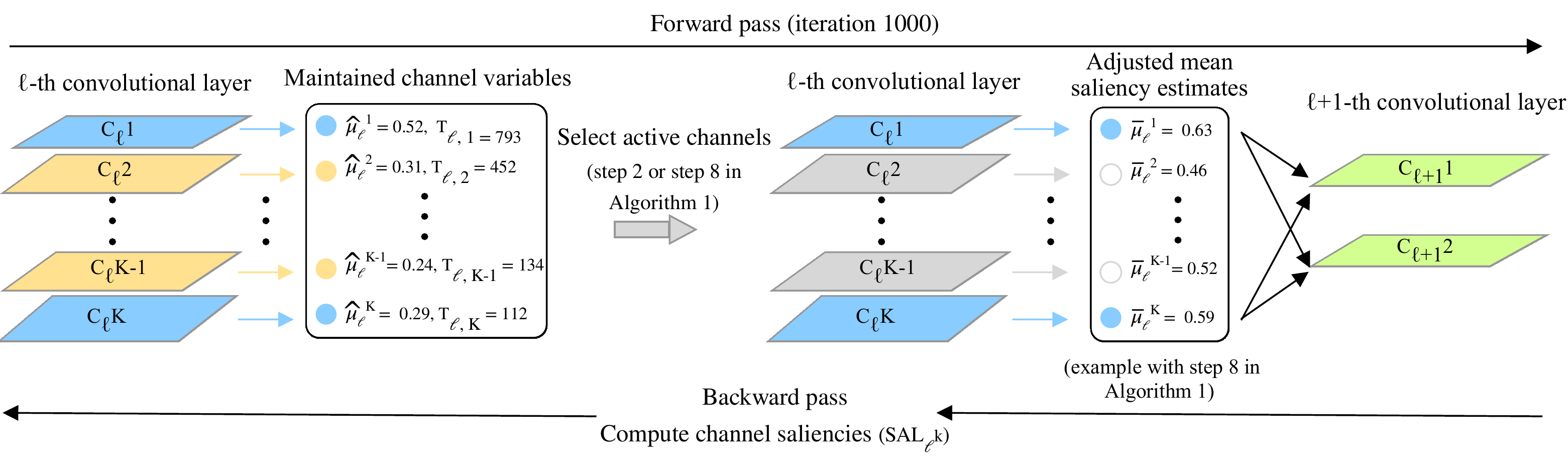}
    \caption{We associate two channel variables with each convolutional channel $C_l^k$: a mean saliency estimate, $\hat{\mu}_l^k$, and a variable which stores the number of times the channel has been activated in all training steps so far, $T_l^k$. Our framework selects a number of channels to activate according to the  combinatorial upper confidence bound algorithm. After initializing the channel variables via randomly sampling channels for a number of training steps, we use $\hat{\mu}_l^k$ and $T_l^k$ to calculate the adjusted mean saliency estimates $\overline{\mu}_l^k$. The channels with the highest adjusted saliency estimates will be activated (in blue colour), whereas the rest (yellow) will be pruned away (grayed out) in the current training step.}
    \label{fig:method_overview}
\end{figure}

We consider a supervised learning problem with a set of training examples $\mathcal{D} = \{ \textbf{X} = \{\textbf{x}_{\texttt{1}}, \textbf{x}_{\texttt{2}}, \ldots, \textbf{x}_{\texttt{N}}  \}, \textbf{Y} = \{ y_{\texttt{1}}, y_{\texttt{2}}, \ldots, y_{\texttt{N}} \} \}$, where $\textbf{x}$ and $y$ represent an input and a label, respectively. Given a CNN model with $L$ convolutional layers, let each layer $l\in 1 \ldots L$ comprise $K_l$ channels, $C_l^k$, where $k \in 1 \ldots K_l$ is the channel index.
In each training step $t$, we 1) sample a batch of $B$ data samples (\textbf{x}\textsuperscript{1:$B$}, \textbf{y}\textsuperscript{1:$B$}); 2) select and activate a subset $S$ of convolutional channels (see Figure~\ref{fig:method_overview}); 3) run a forward and a backward pass on the ``thin'' network, that is only on the active channels; and 4) observe the revealed saliency estimates (SAL$_{l, t}^k$) of the activated channels. The saliency metric we employ was proposed by Molchanov \emph{et al.}~\cite{Molchanov} and approximates the change in loss incurred from removing a particular channel. We maintain an empirical channel saliency mean $\hat{\mu}_l^k$ as well as the number of times $T_l^k$ each channel has been activated in all training steps so far. More precisely, if channel $C_l^k$ has been activated  $T_l^k$ times by the end of training step $t$, then the value of $\hat{\mu}_l^k$ at the end of training step $t$ is ($\sum_{1}^{t}SAL_{l, t}^k$)/$T_l^k$. In addition, it is assumed that the cardinality of $S$, that is the number of channels to be activated in each training step, is predefined and set beforehand. For instance, we might decide to only activate and execute 20$\%$ of all channels at run-time, which would be 1100 channels for the VGG-19 network as an example. After training the network for a given number of iterations, we fix the subset of active channels by selecting the top percentile according to their global rank of mean saliency estimates across all layers, and fine-tune the model. This final fine-tuning stage can be viewed as operating solely in exploitation mode. 

\subsection{Dynamic channel selection mechanism}
Our methodology for channel selection is based on the combinatorial upper confidence bound algorithm (CUCB) \cite{CUCB}, which is a general framework for the combinatorial multi-armed bandit problem (CMAB). In the CMAB setting we are given a system of arms (or resources/ machines) each having unknown utility, that is an unknown distribution of reward with an unknown mean. The task is to select a combination of arms so as to minimize the difference in total expected reward between always playing the optimal super-arm (the combination of arms), and playing super-arms according to the CUCB algorithm. Our modified version of the CUCB framework applied to dynamic channel selection is summarized in Algorithm~\ref{algo}. In our version of the algorithm each network channel can be viewed as an arm, whereas a super-arm constitutes the subset of channels $S$ to be activated at each training step.

Algorithm~\ref{algo} can be loosely divided in two stages: firstly, an initialization round of exploring the saliencies of the channels in the network, and a second stage where we start exploiting and refining the initial saliency estimates to guide the channel selection procedure. The initialization stage comprises steps 1 and 2 in Algorithm~\ref{algo}, whereas stage 2 comprises steps 4 to 11. In the first stage of the algorithm we iterate over all channels in the network, and for each channel $C_l^k$ we activate an arbitrary set of channels S, such that $C_l^k \in S$. We then run the ``thin'' network on a batch of training examples (\textbf{x}\textsuperscript{1:$B$}, $y$\textsuperscript{1:$B$}), and observe the saliency estimates SAL$_l^k$ of the activated channels. This first stage is required to initialize the channel variables $\hat{\mu}_l^k$ and $T_l^k$, such that we can use them to drive the dynamic channel selection mechanism. The second stage of our framework differs in how the active channels are selected. Instead of randomly sampling an arbitrary set of channels, we calculate the \textit{adjusted} channel saliencies $\overline{\mu}_l^k$, and activate the channels with the highest adjusted saliencies at each training step. Specifically, the adjusted channel saliency $\overline{\mu}_l^k$ for $C_l^k$ which has been activated $T_l^k$ times by the end of training step $t$ is $\hat{\mu}_l^k + \sqrt{\frac{3\ln{t}}{2T_l^k}}$. 


\begin{algorithm}
\caption{The CUCB algorithm applied to selective neural network channel execution}\label{algo}
\begin{algorithmic}[1]
\State {For each channel $k$ in each convolutional layer $l$ maintain: 1) $T_l^k$ as the total number of times the particular channel has been activated so far; 2) $\hat{\mu}_l^k$  as the mean of all saliency estimates observed so far.}
\State {For each channel $k$ in each convolutional layer $l$, activate an arbitrary set of channels $S$, such that $C_l^k \in S$, run forward and backward passes through ``thin'' network, and update $T_l^k$ and $\hat{\mu}_l^k$.}
\State $t \gets \text{number of convolutional channels in network}$
\For{\text{j=1\ldots J}}
        \For{batch (\textbf{x}\textsuperscript{1:$B$}, $y$\textsuperscript{1:$B$}) \text{in} $\mathcal{D}$ }
            \State{$t = t + 1$}
            \State{For each channel $C_l^k$, set $\overline{\mu}_l^k = \hat{\mu}_l^k + \sqrt{\frac{3\ln{t}}{2T_l^k}}$}
            \State {$S \gets \text{top percentile of channels according to } \overline{\mu}_l^k$}
            \State{Activate channels in S}
            \State{Run forward and backward passes through ``thin'' network}
            \State{Update all $T_l^k$ and $\hat{\mu}_l^k$}
        \EndFor
\EndFor
\State {$S \gets \text{top percentile of channels according to } \hat{\mu}_l^k$}
\State{Activate channels in S}
\State{Fine-tune final ``thin'' network}

\end{algorithmic}
\end{algorithm}

In our implementation of dynamic channel execution we use masking, i.e. multiplying the output of inactive channels by zero, to sparsify the full networks. An alternative approach would be to instantiate a new compact model only consisting of active channels and copy the corresponding weights from the original sparse network. This would have to be performed at each training step to benefit from a lower memory footprint and floating-point operations at run-time. 

\subsection{Estimating channel saliency}
Our dynamic channel selection framework requires the estimation of channel saliency, or the contribution of each active channel to the overall network performance. We need a saliency metric which enables us to rank the filters of the entire network globally, that is across layers. Molchanov \emph{et al.} ~\cite{Molchanov} propose a pruning framework which leverages a first-order Taylor approximation for global channel ranking, whereas Theis \emph{et al.} ~\cite{Theis} use Fisher information to achieve kernel ranking across layers~\cite{Figurnov}. Our channel ranking approach is based on Molchanov \emph{et al.}~\cite{Molchanov} although both methods would be applicable. 

The intuition behind the approach is approximating the change in the loss function from removing a particular channel, which was active at training step $t$, via a first-order Taylor expansion. Let $h_l^k$ be the feature map produced by the channel $C_l^k$. Also, let $\mathcal{\textbf{b}}$ denote a single batch (\textbf{x}\textsuperscript{1:$B$}, $y$\textsuperscript{1:$B$}) from the training data $\mathcal{D}$. Then $L(\mathcal{\textbf{b}},h_l^k)$ will denote the loss of the network when channel $C_l^k$ is active. On the other hand, we will use the notation $L(\mathcal{\textbf{b}}, h_l^k = 0)$ to denote the loss of the network had $C_l^k$ been inactive at the given training step.

\begin{equation}
\left| \Delta L (h_l^k) \right| = \left| L (\mathcal{\textbf{b}}, h_l^k = 0) -  L (\mathcal{\textbf{b}}, h_l^k) \right| = \hat{SAL}_{l,t}^k
\label{eq: Eq. 1}
\end{equation}
Intuitively, our salience metric considers channels which incur a high increase in loss when deactivated as more important. After a first-order approximation around the point $h_l^k = 0$, Eq.~\ref{eq: Eq. 1} can be shown to be equal to:

\begin{equation}
\hat{SAL}_{l,t}^k = \left| \frac{1}{M} \sum_{m = 1}^{M} \frac{\delta L}{\delta h_{l,m}^k} h_{l,m}^k \right|,
\label{eq: Eq. 2}
\end{equation}
where  $M$ is the length of the vectorized feature map. Eq.~\ref{eq: Eq. 2} assumes independence between channel parameters whereas in reality they are inter-connected. The computation of Eq.~\ref{eq: Eq. 2} is easy in practice as it only requires the first derivative of the loss w.r.t each feature map element, i.e. $\frac{\delta L}{\delta h_{l,m}^k}$, which is computed during backpropagation. In order to compare the saliencies of the activated channels after each training step, we need to normalize the raw values as their scale varies across layers. We use $l_2$-normalization before ranking the saliencies:

\begin{equation}
SAL_{l,t}^k = \frac{\hat{SAL}_{l,t}^k}{\sqrt{\sum_{j}(\hat{SAL}_{l,t}^j)^2 }}
\label{eq: Eq. 3}
\end{equation}

The saliency metric is used to update the mean saliency estimates $\hat{\mu}_l^k$ of all active channels.

\section{Experiments}
In this section we empirically demonstrate the effectiveness of our framework for dynamic channel selection during training. We conduct all experiments in PyTorch~\cite{pytorch}.

\begin{table}[ht!]
\centering
{\small (a) Test Accuracy on CIFAR-10}
\begin{subtable}
\centering
        \resizebox{0.9\textwidth}{!}{%
         \begin{tabular}{c | c c c c c} 
         \hline
         Model & Test accuracy & Parameters & $\%$ Baseline & FLOPs & $\%$ Baseline \\ [0.5ex] 
         \hline
         VGG-19(Baseline) & $92.86\%$ & $20.035\times10^6$ & - &$3.98\times10^8$ & -  \\ 
         VGG-19($70\%$ channels) & \B 93.11$\%$ & $9.57\times10^6$ & $48\%$ &$2.84\times10^8$ & $71\%$  \\
         VGG-19($40\%$ channels) & $92.02\%$ & $2.89\times10^6$ & $14\%$ &$1.32\times10^8$ & $33\%$  \\ 
         \hline
         ResNet-50(Baseline) & 92.94$\%$ & $23.52\times10^6$ & - &$0.85\times10^8$ & -  \\ 
         ResNet-50($90\%$ channels) &  \B93.52$\%$ & $20.96\times10^6$ & $89\%$ &$0.74\times10^8$ & $87\%$  \\ 
         ResNet-50($30\%$ channels) & $92.93\%$ & $5.48\times10^6$ & $23\%$ &$0.3\times10^8$ & $35\%$  \\ 
         \hline 
         DenseNet-121(Baseline) & $93.22\%$ & $6.96\times10^6$ & - &$0.59\times10^8$ & -  \\ 
         DenseNet-121($80\%$ channels) & \B 93.73$\%$ & $5.14\times10^6$ & $74\%$ &$0.44\times10^8$ & $75\%$  \\ 
         DenseNet-121($40\%$ channels) & $92.82\%$ & $1.37\times10^6$ & $20\%$ &$0.13\times10^8$ & $22\%$  \\ 
         \hline
         \end{tabular}
         }

\end{subtable}

\bigskip
{\small (b) Test Accuracy on CIFAR-100}
\begin{subtable}
\centering
    \resizebox{0.9\textwidth}{!}{%
     \begin{tabular}{c | c c c c c} 
     \hline
     Model & Test accuracy & Parameters & $\%$ Baseline & FLOPs & $\%$ Baseline \\ [0.5ex] 
     \hline
     VGG-19(Baseline) & $68.37\%$ & $20.08\times10^6$ & - &$4\times10^8$ & -  \\ 
     VGG-19($90\%$ channels) & \B70.31$\%$ & $16.16\times10^6$ & $80\%$ &3.77$\times10^8$ & $94\%$  \\
     VGG-19($40\%$ channels) & $66.72\%$ & $2.85\times10^6$ & $14\%$ &$1.48\times10^8$ & $37\%$  \\ 
     \hline
     ResNet-50(Baseline) & $72.88\%$ & $23.7\times10^6$ & - &$0.86\times10^8$ & -  \\ 
     ResNet-50($90\%$ channels) & \B73.04$\%$ & $21.6\times10^6$ & $91\%$ &$0.75\times10^8$ & $87\%$  \\ 
     ResNet-50($40\%$ channels) & $70.34\%$ & $8.45\times10^6$ & $36\%$ &$0.4\times10^8$ & $47\%$  \\ 
     \hline
     DenseNet-121(Baseline) & 74.08$\%$ & $7.05\times10^6$ & - &$0.6\times10^8$ & -  \\ 
     DenseNet-121($90\%$ channels) & \B75.07$\%$ & $6.21\times10^6$ & $88\%$ &$0.52\times10^8$ & $87\%$  \\ 
     DenseNet-121($40\%$ channels) & $70.98\%$ & $1.4\times10^6$ & $20\%$ &$0.14\times10^8$ & $23\%$  \\ 
     \hline
     \end{tabular}
     }
\end{subtable}

\bigskip
{\small (a) Test Accuracy on SVHN}
\begin{subtable}
\centering
    \resizebox{0.9\textwidth}{!}{%
     \begin{tabular}{c | c c c c c} 
     \hline
     Model & Test accuracy & Parameters & $\%$ Baseline & FLOPs & $\%$ Baseline \\ [0.5ex] 
     \hline
     VGG-19(Baseline) & $95.57\%$ & $20.035\times10^6$ & - &$3.98\times10^8$ & -  \\ 
     VGG-19($80\%$ channels) & \B96.04$\%$ & $12.64\times10^6$ & $63\%$ &$3.22\times10^8$ & $81\%$  \\
     VGG-19($40\%$ channels) & $95.52\%$ & $2.81\times10^6$ & $14\%$ &$1.43\times10^8$ & $36\%$  \\ 
     \hline
     ResNet-50(Baseline) & $92.34\%$ & $23.52\times10^6$ & - &$0.85\times10^8$ & -  \\ 
     ResNet-50($30\%$ channels) & \B94.39$\%$ & $5.49\times10^6$ & $23\%$ &$0.27\times10^8$ & $32\%$  \\ 
     \hline
     DenseNet-121(Baseline) & $94.07\%$ & $6.96\times10^6$ & - &$0.59\times10^8$ & -  \\ 
     DenseNet-121($90\%$ channels) & \B94.8$\%$ & $6.11\times10^6$ & $88\%$ &$0.51\times10^8$ & $86\%$  \\ 
     DenseNet-121($40\%$ channels) & $94.31\%$ & $1.36\times10^6$ & $20\%$ &$14.4\times10^8$ & $24\%$  \\ 
     \hline
    \end{tabular}}
\end{subtable}
\caption{"Baseline" denotes training the original full networks without dynamic channel selection. In column-1 "40 channels" denotes a model which was trained using our proposed framework with $40\%$ of the channels activated at each training step, etc. The test accuracy we report is after fine-tuning the final ``thin'' network. The ratio of parameters and FLOPs w.r.t the baseline model ($\%$ Baseline) utilized are shown in column-$4\&6$. }
\label{tab:big_table}
\end{table}


\subsection{Datasets}
We evaluate the performance of our method on three datasets: CIFAR-10, CIFAR-100~\cite{cifar} and Street View House Number (SVHN) ~\cite{svhn}. Both CIFAR datasets consist of coloured natural images with resolution 32x32, and comprise 50,000 training examples and 10,000 testing examples. CIFAR-10 is drawn from 10 classes, whereas CIFAR-100 from 100 classes. The SVHN dataset consists of 32x32 coloured digit images. We use all 604,388 training images from SVHN and validate on the test set of 26,032 images. We normalize the input data using the channel means and standard deviations. The data augmentation scheme we use consists of padding the images with 4 on each border, then randomly cropping and horizontally flipping a 32x32 patch.

\subsection{Network models}
We evaluate our method for dynamic channel selection on three network architectures: VGGNet~\cite{vggnet}, ResNet~\cite{resnet} and DenseNet~\cite{densenet}. We use a VGG-19 architecture of type "VGG-E", however we use only a single fully connected layer (16conv + 1FC)~\cite{vgg_github}. For ResNet we employ a 50-layer architecture with a bottleneck structure (ResNet-50). For DenseNet we use a 121-layer architecture with a growth rate of 32 (DenseNet-121). We use batch normalization on all network models and remove all dropout layers. We do not apply the dynamic channel selection procedure on fully connected layers as they comprise only the final classification layer in our models.

\subsection{Training and Fine-tuning}
All network models are trained using SGD with Nesterov momentum of 0.9 without dampening. The minibatch size we use is 64 for all datasets. On the two CIFAR datasets we train for 160 epochs, and on the SVHN dataset for 20 epochs. The initial learning rate is set at 0.1 and is divided by 10 at 50$\%$ and 75$\%$ of the training epochs. We also use a weight decay of $10^{-4}$. We adopt the weight initialization introduced by~\cite{he_init}. After training following the proposed framework, we fix the subset of active channels by selecting the most salient ones across all layers. We then build a compact model consisting only of active channels and copy the corresponding weights. This compact model is fine-tuned by repeating the training procedure. We evaluate the floating-point operations (FLOPs) and the number of parameters of our compact networks using~\cite{thop}. 

\subsection{Results}

\begin{figure}[!ht]
\centering
{\small (a) Test accuracy on CIFAR-10\par}{
\resizebox{0.99\textwidth}{!}{
\begin{tabular}{ccc}
{\includegraphics{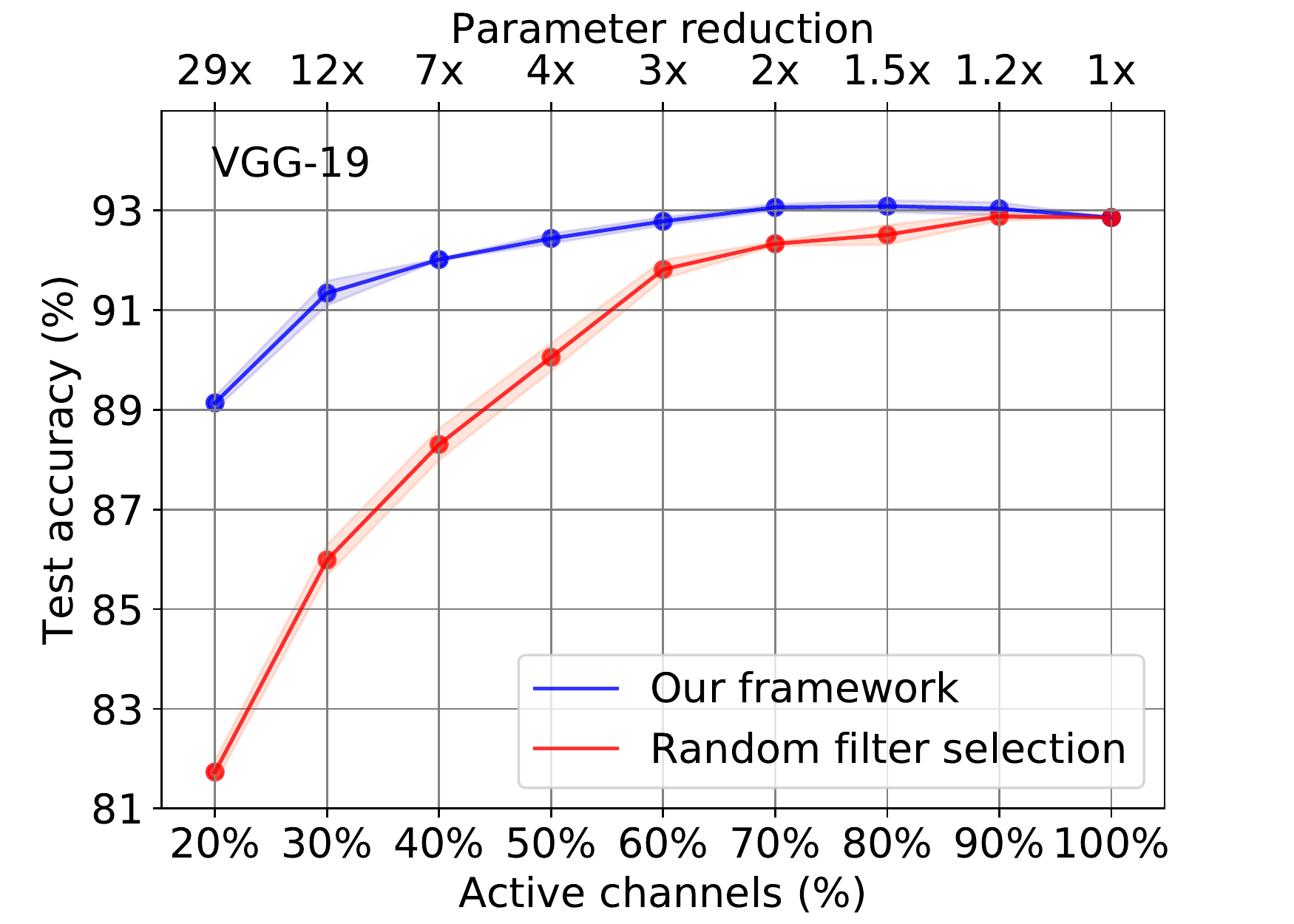}}&
{\includegraphics{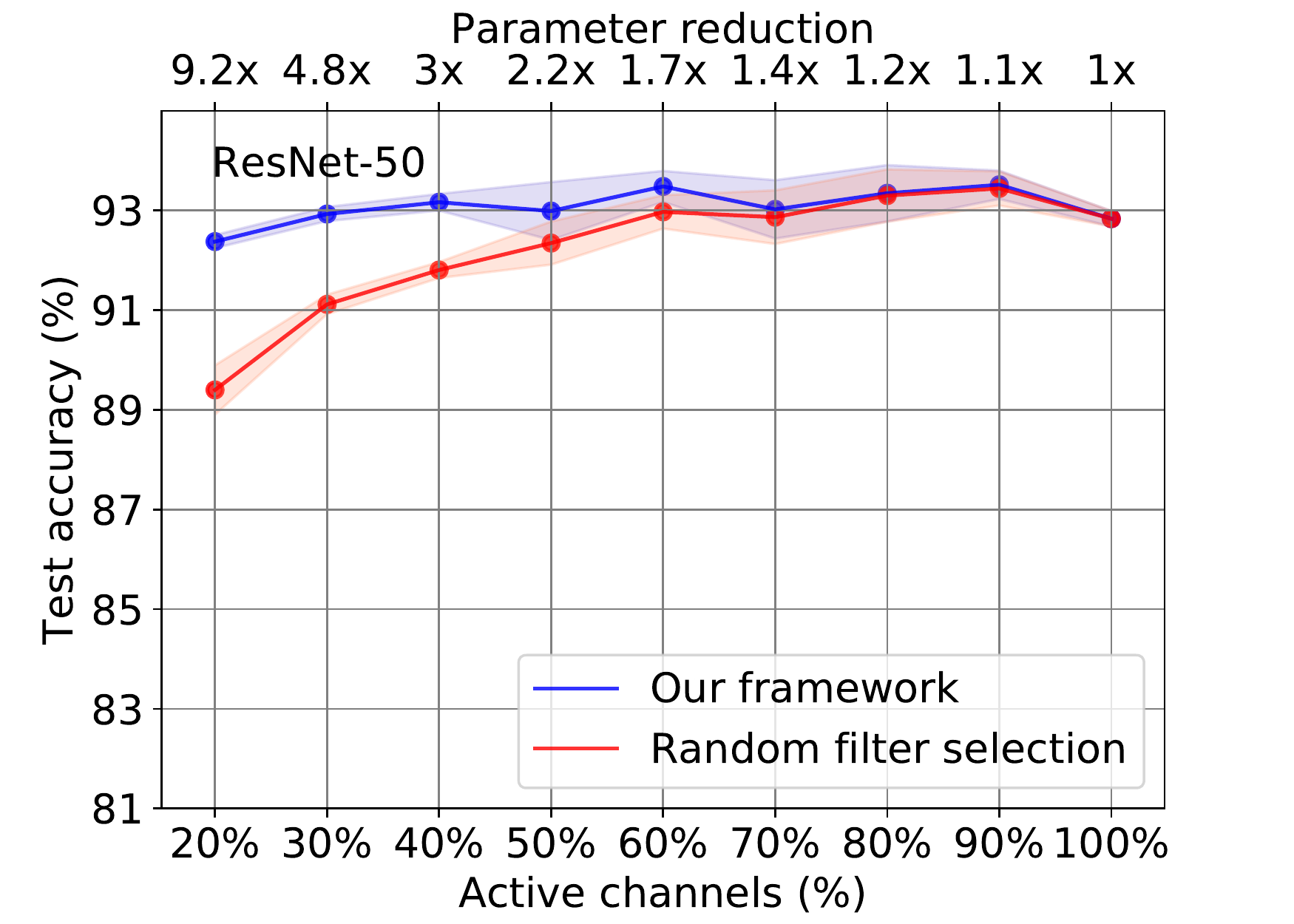}}&
{\includegraphics{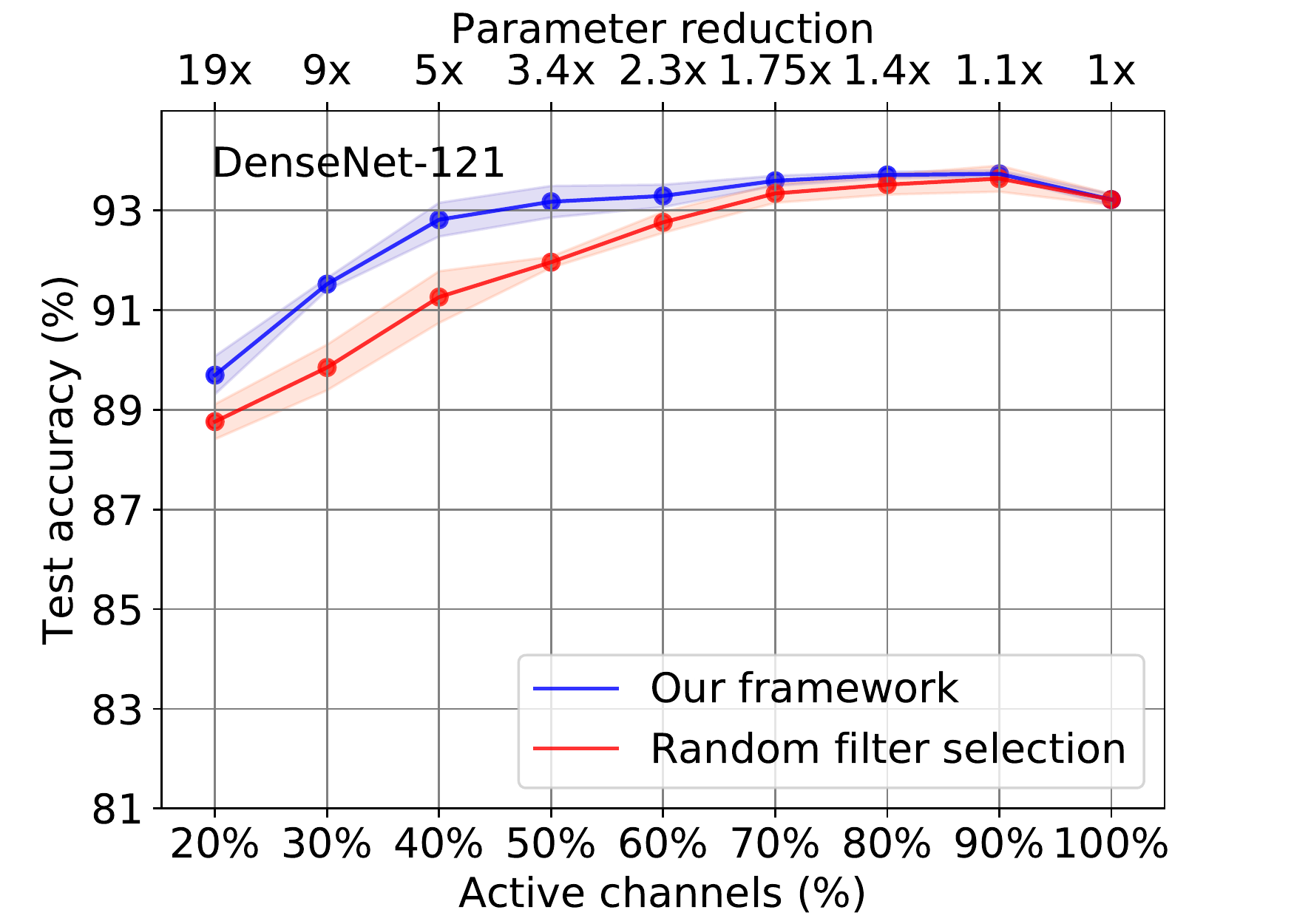}}
\end{tabular} 
} }
\centering
{\small (b) Test accuracy on CIFAR-100\par}{
\resizebox{0.99\textwidth}{!}{
\begin{tabular}{ccc}
{\includegraphics{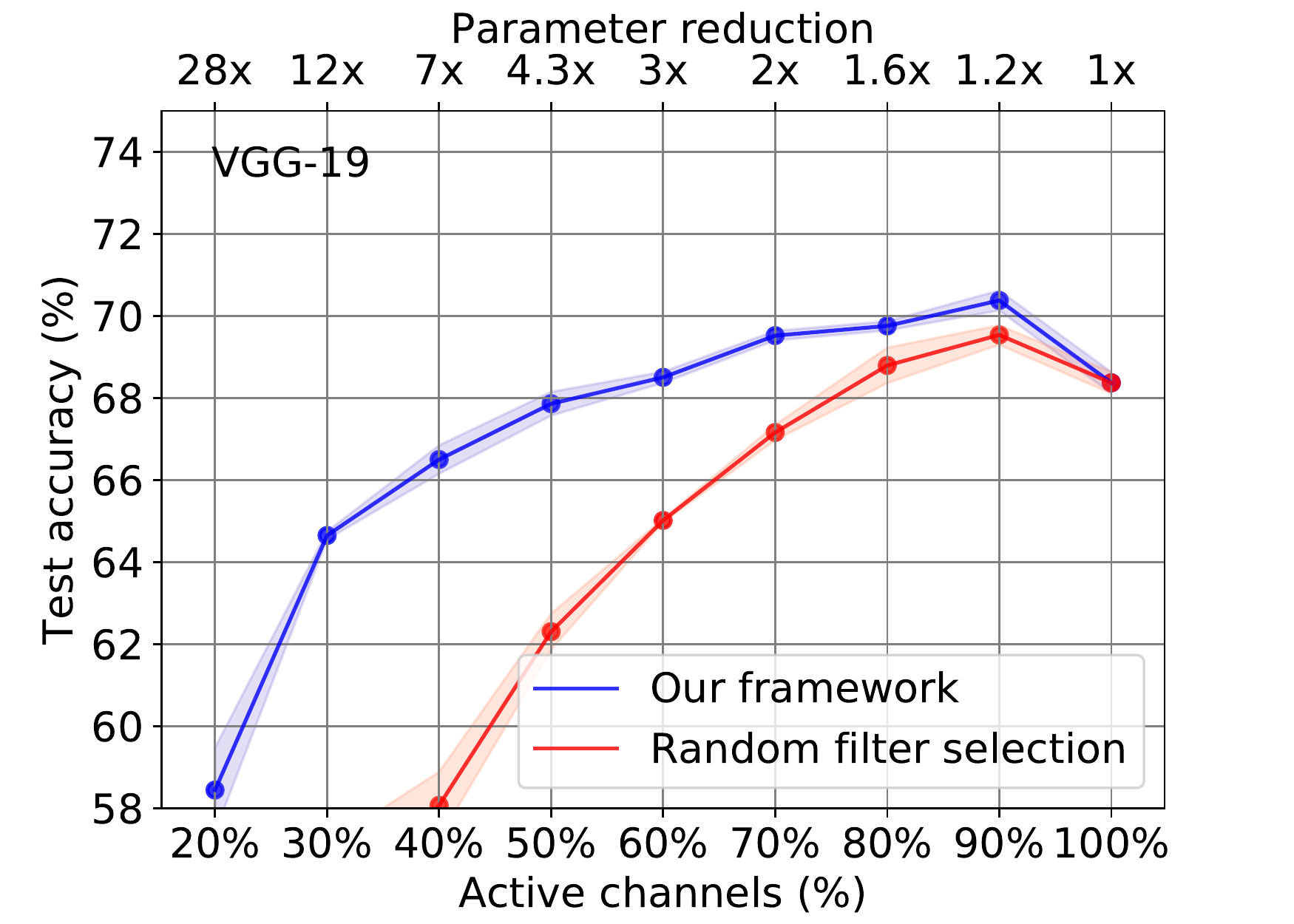}}&
{\includegraphics{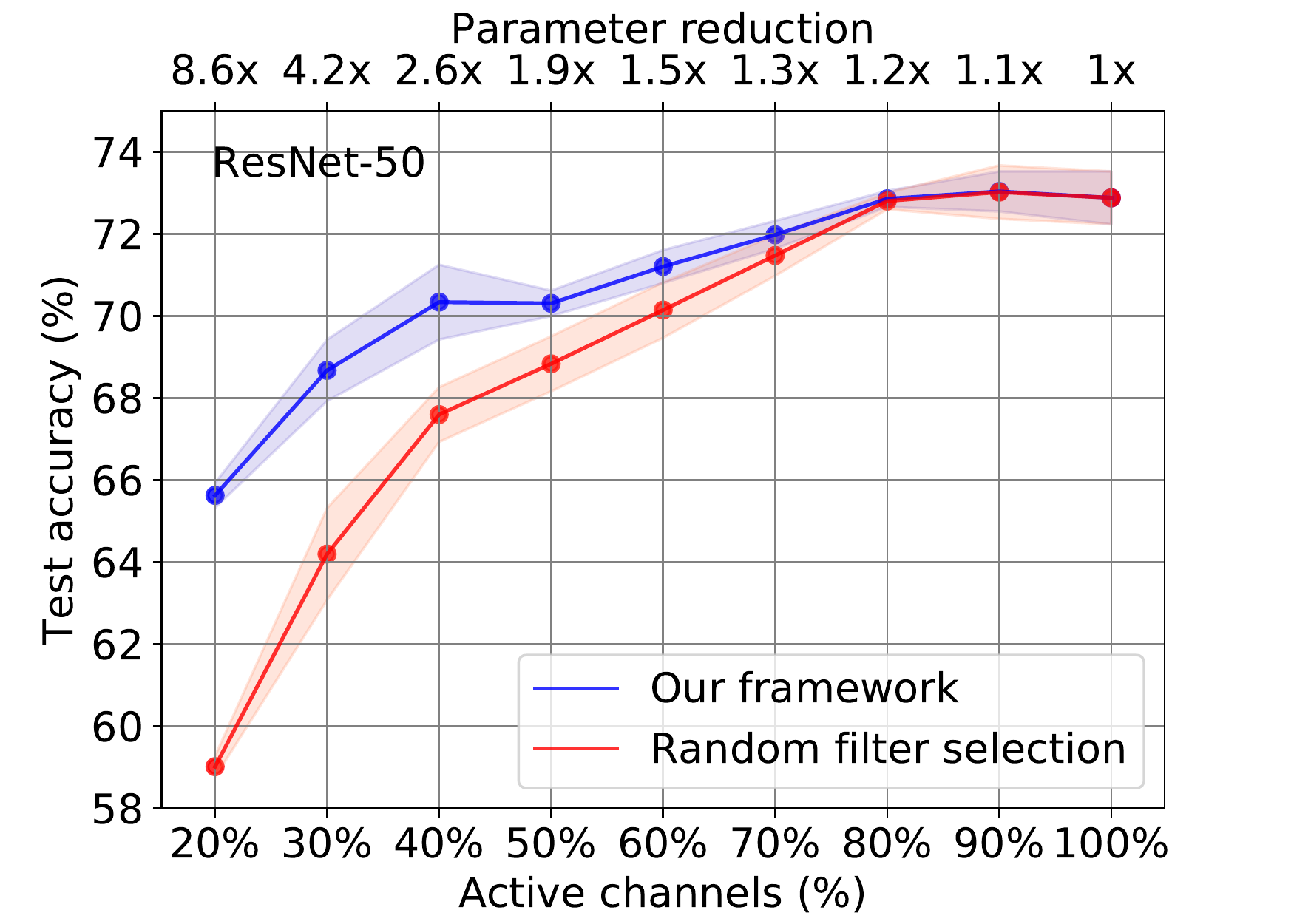}}&
{\includegraphics{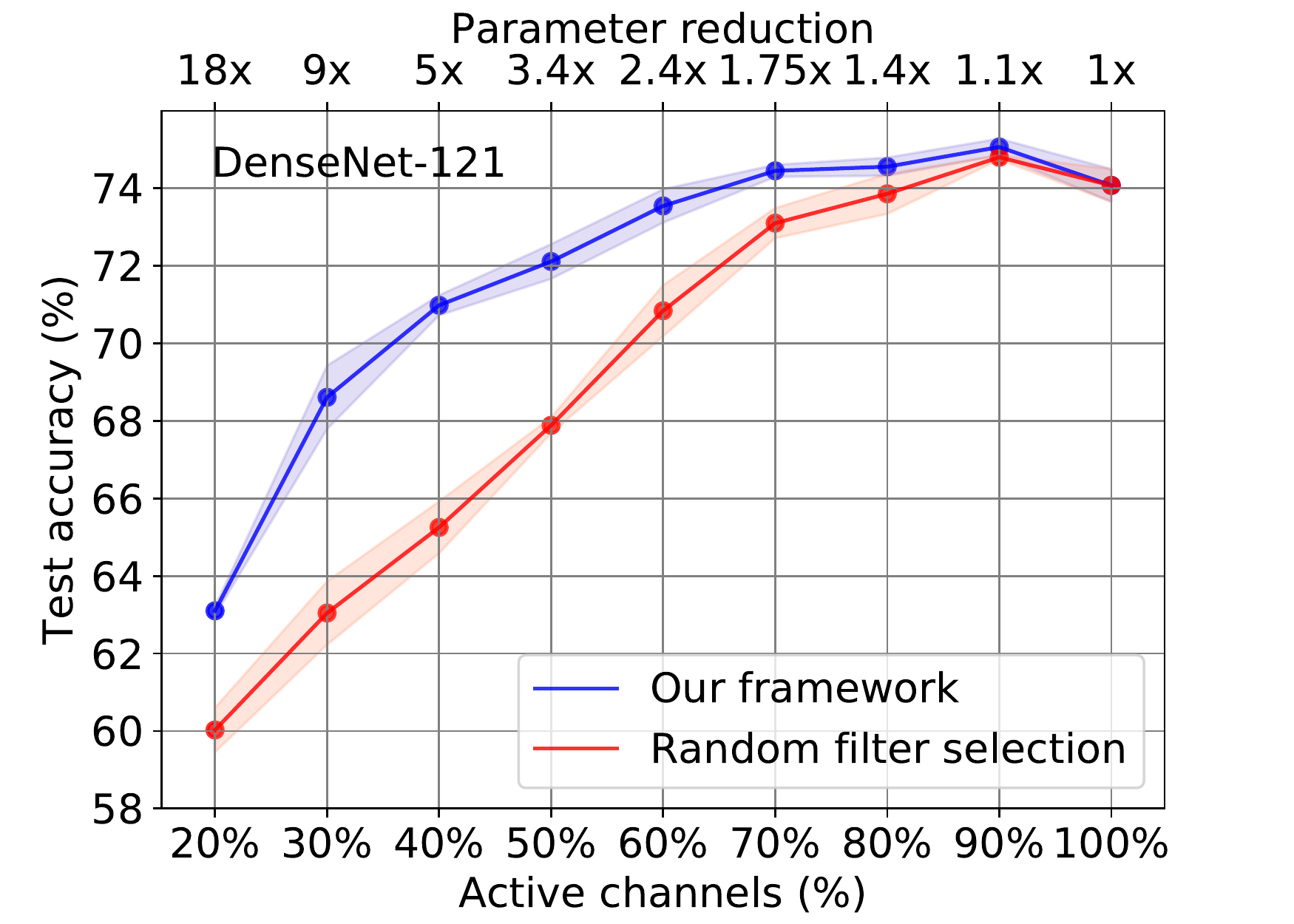}}
\end{tabular} 
} }
\centering
{\small (c) Test accuracy on SVHN\par}{
\resizebox{0.99\textwidth}{!}{
\begin{tabular}{ccc}
\includegraphics{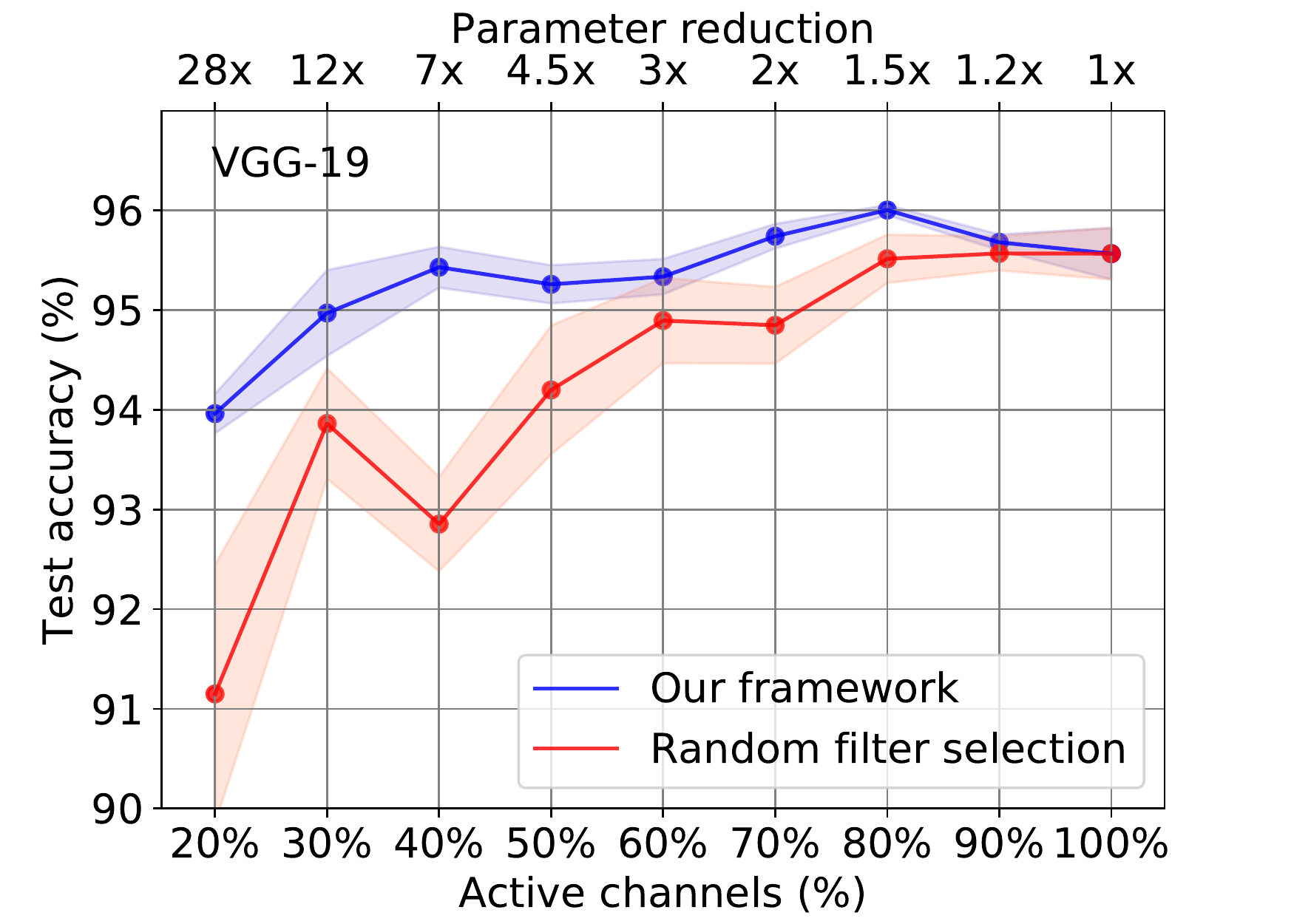}&
{\includegraphics{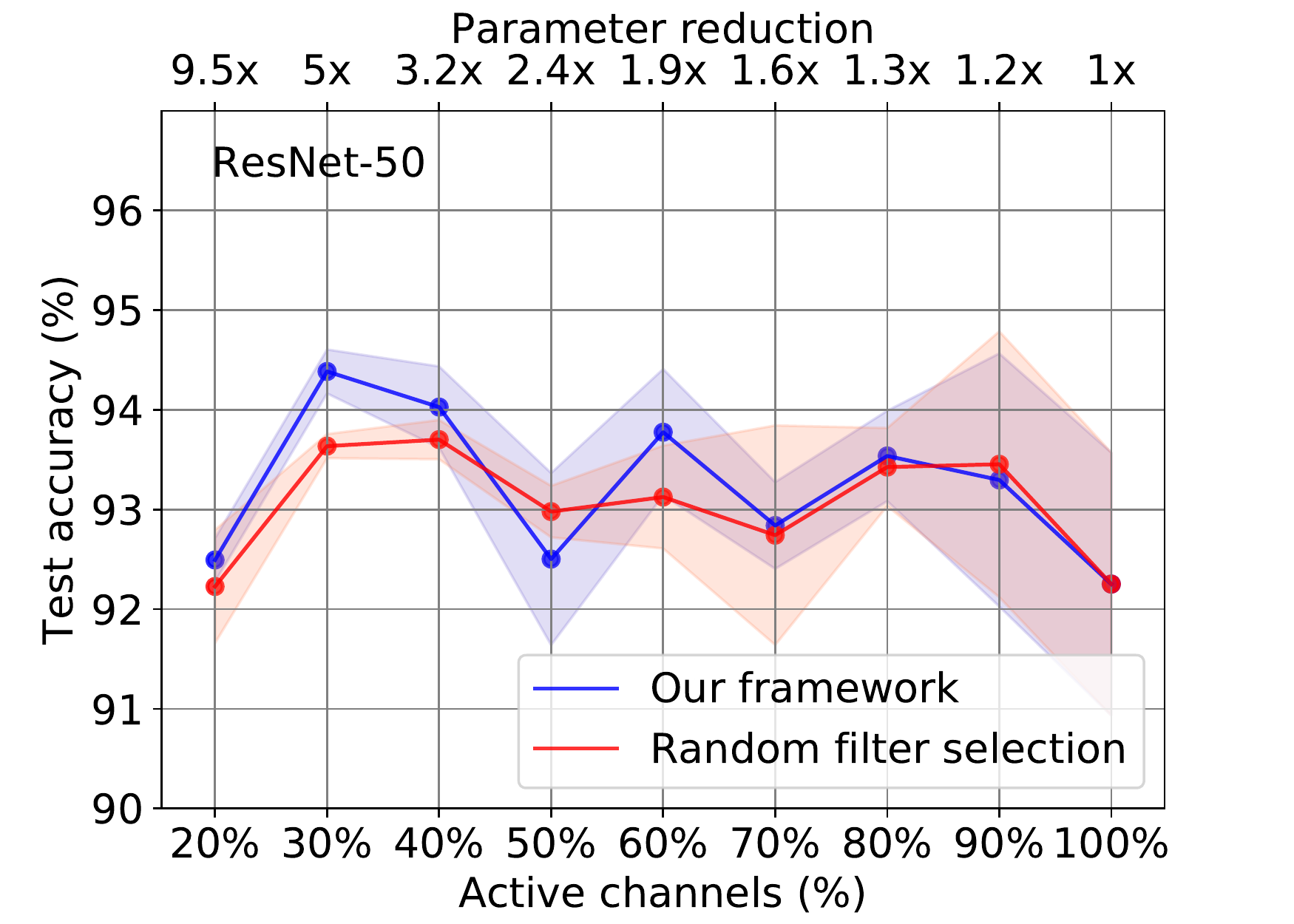}}&
{\includegraphics{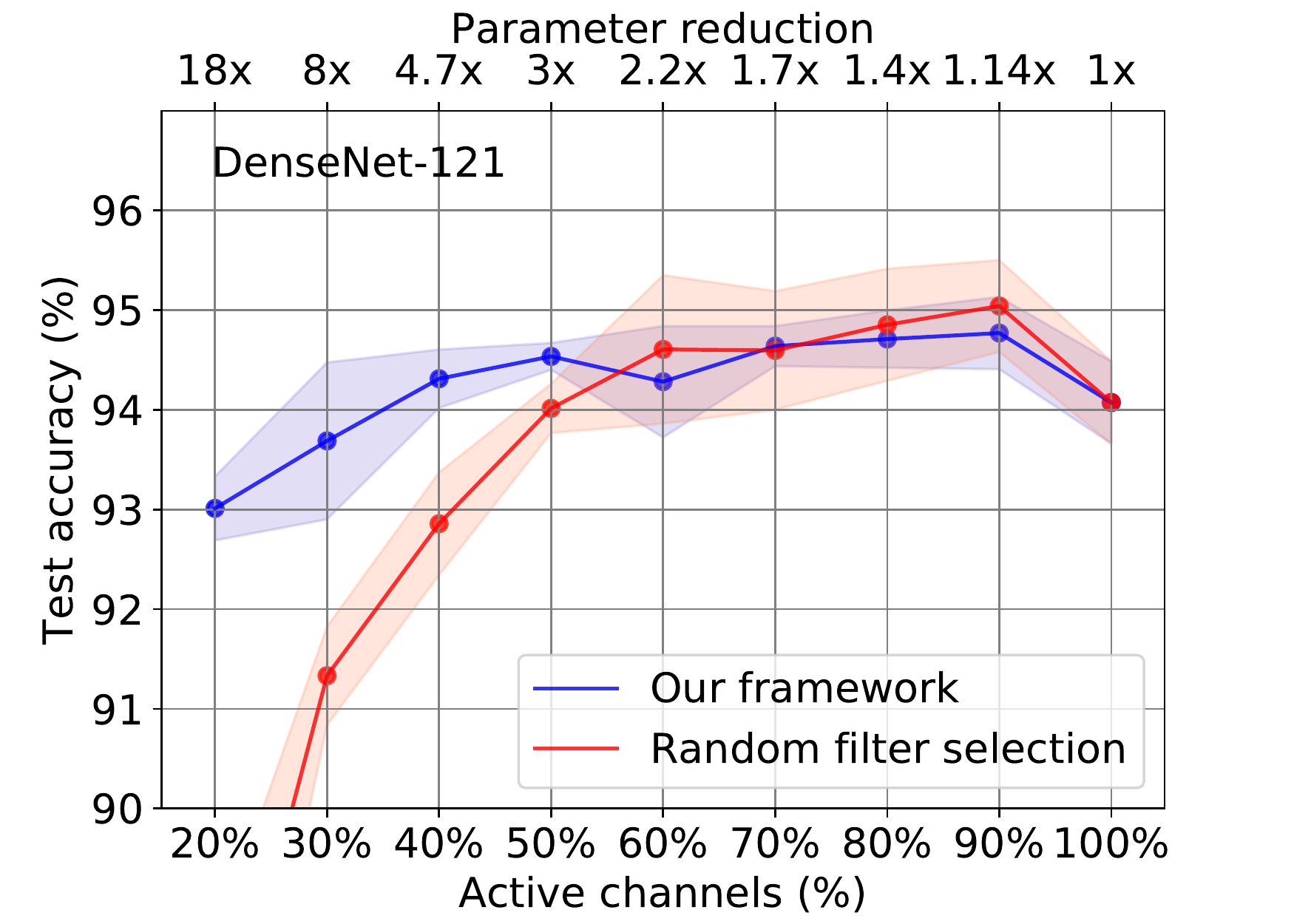}}
\end{tabular} 
} }

\caption{Comparing our proposed framework to random channel selection. The coloured areas denote standard deviation. All experiments were run three times. In general, our CUCB-based framework outperforms random channel selection.}
\label{fig:pareto}
\end{figure}

We evaluate the dynamic channel selection framework on all combinations of datasets and network architectures. We experiment with a ratio of active channels between 20$\%$ to 100$\%$ in 10$\%$ increments. We perform all experiments three times and report averaged results. Table~\ref{tab:big_table} shows the highest test accuracies achieved by our models in \textbf{boldface}. \par \vspace{1em}

\textbf{Regularization effect.} From table~\ref{tab:big_table} we can observe that for all datasets and network models the highest test accuracy is achieved when the percentage of active channels is typically between $70\%$-$90\%$. The only exception is ResNet-50 evaluated on the SVHN dataset which achieves peak classification when $30\%$ of the channels are active. We hypothesize that the increase in accuracy is due to the regularization effect of the dynamic channel selection procedure which could be viewed as feature selection applied on the hidden layers. A similar phenomenon is observed in the pruning frameworks~\cite{net_slimming}, where L1 regularization is imposed on channels instead, as well as~\cite{bmvc_pruning}, ~\cite{structured_sparsity}. Finally, it is noteworthy to mention that our framework outperforms Dynamic Deep Neural Networks ~\cite{michigan_dnn} in terms of relative drop in accuracy compared to baseline for ResNet evaluated on CIFAR-10. \cite{michigan_dnn} observe up to $7\%$ drop in accuracy whereas our framework maintains baseline performance even with $20\%$ active channels (see Figure~\ref{fig:pareto}). \par \vspace{1em}

\textbf{Parameter and FLOP reduction.} The main goal of our dynamic channel execution framework is to lower the demands for training parameter-efficient networks. Since the best performing models use active channels between $70\%$-$90\%$, their parameter count and FLOPs are also lower compared to baseline models. For example, on CIFAR-10 the best performing VGG model achieves $2\times$ parameter reduction. DenseNet-121 and especially ResNet-50 are unable to achieve such parameter-efficiency owing to their bottleneck architecture. More specifically, the skip connections in ResNet-50 based on the addition operation require that certain convolutional layers have the same number of output channels. In table~\ref{tab:big_table} we have also shown results from smaller networks with $30\%$-$40\%$ active channels (last row of each model). We observe comparable accuracy to baseline for these very compact networks on the CIFAR-10 and SVHN datasets with parameter reduction between $3\times$-$7\times$ and FLOPs reduction $2\times$-$5\times$. Moreover, ResNet-50 can achieve over $9\times$ parameter reduction on CIFAR-10 and SVHN (see Figure~\ref{fig:pareto}) while maintaining baseline-level performance. Nevertheless, the small networks perform worse than baseline on the CIFAR-100 dataset ($2\%$-$3\%$ accuracy drop for models with $30\%$-$40\%$ active channels). We conjecture this is because CIFAR-100 contains 100 classes and extra model capacity is required. \par \vspace{1em}

\textbf{Single-stage training.} Instead of following the proposed two-stage process of firstly training a model with the dynamic channel selection framework, and then fine-tuning by repeating the same training procedure on the most salient channels, we attempt an integrated pipeline. We experiment with the VGG-19 network on CIFAR-10 with the same hyperparameter settings as described in Section 3.3. For the first 40 iterations we use the proposed dynamic channel selection framework, then we activate the most salient channels and freeze the network topology for the remainder of the training process. We compare the performance of the single-stage and two-stage training procedures in Table~\ref{tab:comparison_table}. Overall, we observe a degradation in performance for the integrated training procedure with a test accuracy drop between $0.18\%$-$0.86\%$. \par \vspace{1em}
\begin{table}[!ht]
    \centering
    \resizebox{0.9\textwidth}{!}{%
    \begin{tabular}{c | c c c c c c c c} 
    \hline
         Active channels ($\%$) & 20$\%$ & 30$\%$ & 40$\%$ & 50$\%$ & 60$\%$ & 70$\%$ & 80$\%$ & 90$\%$  \\ \hline
         Single-stage test accuracy ($\%$) & 89.4 & 90.98 & 91.15 & 91.74 & 91.96 & 92.2 & 91.91 & 92.24  \\
         Accuracy gain ($\%$) & +0.26 & -0.36 & -0.86 & -0.7 & -0.82 & -0.86 &-0.18 & -0.8 \\ 
        \hline
    \end{tabular}
    }
    \caption{The test accuracy performance for VGG-19 on CIFAR-10 with a single-stage training process. Accuracy gain (- sign means accuracy drop) is compared to the equivalent two-stage training procedure.}
    \label{tab:comparison_table}
\end{table}
\textbf{Results on CUCB vs random channel selection.} We performed experiments where instead of activating the channels with the highest mean estimated saliency prior to fine-tuning (step 12 in Algorith~\ref{algo}), we randomly select a subset of channels to activate (see Figure~\ref{fig:pareto}). It can be observed that randomly selecting channels to fine-tune performs worse in general than our proposed methodology. The difference in performance becomes more significant as the active channels become fewer. The sequential VGG-19 architecture adheres to this behaviour on all datasets. For ResNet-50 and DenseNet-121 the performance of models fine-tuned on randomly selected channels is on occasion on par with our framework. More specifically, on CIFAR-10 when the active channels are $\geq 70\%$, and on SVHN when random channel selection even outperforms our framework on a few instances. We hypothesize our methodology, which is based on the assumption of independence between channels, might be adversely affected by architectures with skip connections across layers, such as ResNet and DenseNet. 


\section{Conclusion}
In this paper we have proposed a novel methodology, inspired by the combinatorial multi-armed bandit problem, to dynamically identify and utilize only the most salient convolutional channels at each training step. As a result, we can limit the memory and computational burden both during training \textit{and} inference. Our method tracks the relative importance of each channel, and at each training step a subset of highly salient channels are activated and executed according to the combinatorial upper confidence bound algorithm. Experimental results on several datasets and state-of-the-art network architectures reveal out framework is able to significantly reduce computational cost (up to $4\times$) and parameter count (up to $9\times$) with little or no loss in accuracy. We also demonstrate our framework consistently performs better than random channel selection. 

\bibliography{arXiv_dyn_sel}
\bibliographystyle{unsrt}  


\end{document}